\documentclass[letterpaper, 10 pt, conference]{ieeeconf}  

\usepackage{graphicx}
\usepackage{multirow}
\usepackage{amsmath}
\usepackage{amssymb}
\usepackage{array}
\usepackage[font={small}]{caption}
\usepackage{subcaption}
\usepackage[dvipsnames]{xcolor}
\usepackage{color,soul}
\usepackage{url}
\usepackage[ruled,vlined]{algorithm2e}
\usepackage{booktabs}

\IEEEoverridecommandlockouts                              

\title{\LARGE \bf
Decentralized Multi-Robot Navigation for Autonomous Surface Vehicles with Distributional Reinforcement Learning \vspace{-3mm}
}

\author{Xi Lin, Yewei Huang, Fanfei Chen, and Brendan Englot
\thanks{Xi Lin, Yewei Huang and Brendan Englot are with the Department of Mechanical Engineering, Stevens Institute of Technology, Hoboken, NJ, USA,
{\tt\small \{xlin26,yhuang85,benglot\}@stevens.edu}.
}%
\thanks{Fanfei Chen is with Aescape, New York, NY, USA.}
}

\begin{document}

\maketitle
\thispagestyle{empty}
\pagestyle{empty}

\begin{abstract}
Collision avoidance algorithms for Autonomous Surface Vehicles (ASV) that follow the Convention on the International Regulations for Preventing Collisions at Sea (COLREGs) have been proposed in recent years. 
However, it may be difficult and unsafe to follow COLREGs in congested waters, where multiple ASVs are navigating in the presence of static obstacles and strong currents, due to the complex interactions.
To address this problem, we propose a decentralized multi-ASV collision avoidance policy based on Distributional Reinforcement Learning, which considers the interactions among ASVs as well as with static obstacles and current flows. 
We evaluate the performance of the proposed Distributional RL based policy against a traditional RL-based policy and two classical methods, Artificial Potential Fields (APF) and Reciprocal Velocity Obstacles (RVO), in simulation experiments, which show that the proposed policy achieves superior performance in navigation safety, while 
requiring minimal travel time and energy. A variant of our framework that automatically adapts its risk sensitivity is also demonstrated to improve ASV safety in highly congested environments.
\end{abstract}

\vspace{-1mm}

\section{Introduction}

Reliable collision avoidance is crucial to the deployment of Autonomous Surface Vehicles (ASV), which is still a challenging problem \cite{vagale2021path}, especially in congested environments.
The Convention on the International Regulations for Preventing Collisions at Sea (COLREGs) \cite{international1972convention} defines rules for collision avoidance between vessels, which have been used to develop obstacle avoidance solutions in multi-ASV  navigation scenarios \cite{kuwata2013safe, kufoalor2018proactive, zhao2019colregs, cho2020efficient}.   
However, following COLREGs may cause conflicting actions when the roles of each ego vehicle with respect to different neighboring vehicles are in conflict, especially in congested waters \cite{jeong2022motion}.
In addition to the risk of collision with other vehicles, static obstacles that exist in the marine environment, such as buoys or rocks, may pose threats to navigation safety.
To avoid collision with them, an ASV may need to maneuver in a way that does not comply with COLREGs, which primarily consider the relationship to nearby vehicles.
Additionally, the motion of marine autonomous vehicles is affected by current flows \cite{lolla2015path}, thus following COLREGs may not always be safe when navigating in strong current.
The objective of this work is to develop a safe decentralized ASV collision avoidance policy for the multi-ASV navigation problem in congested marine environments where unknown static obstacles and currents exist. 
\begin{figure}
    \centering
    \includegraphics[width=\linewidth]{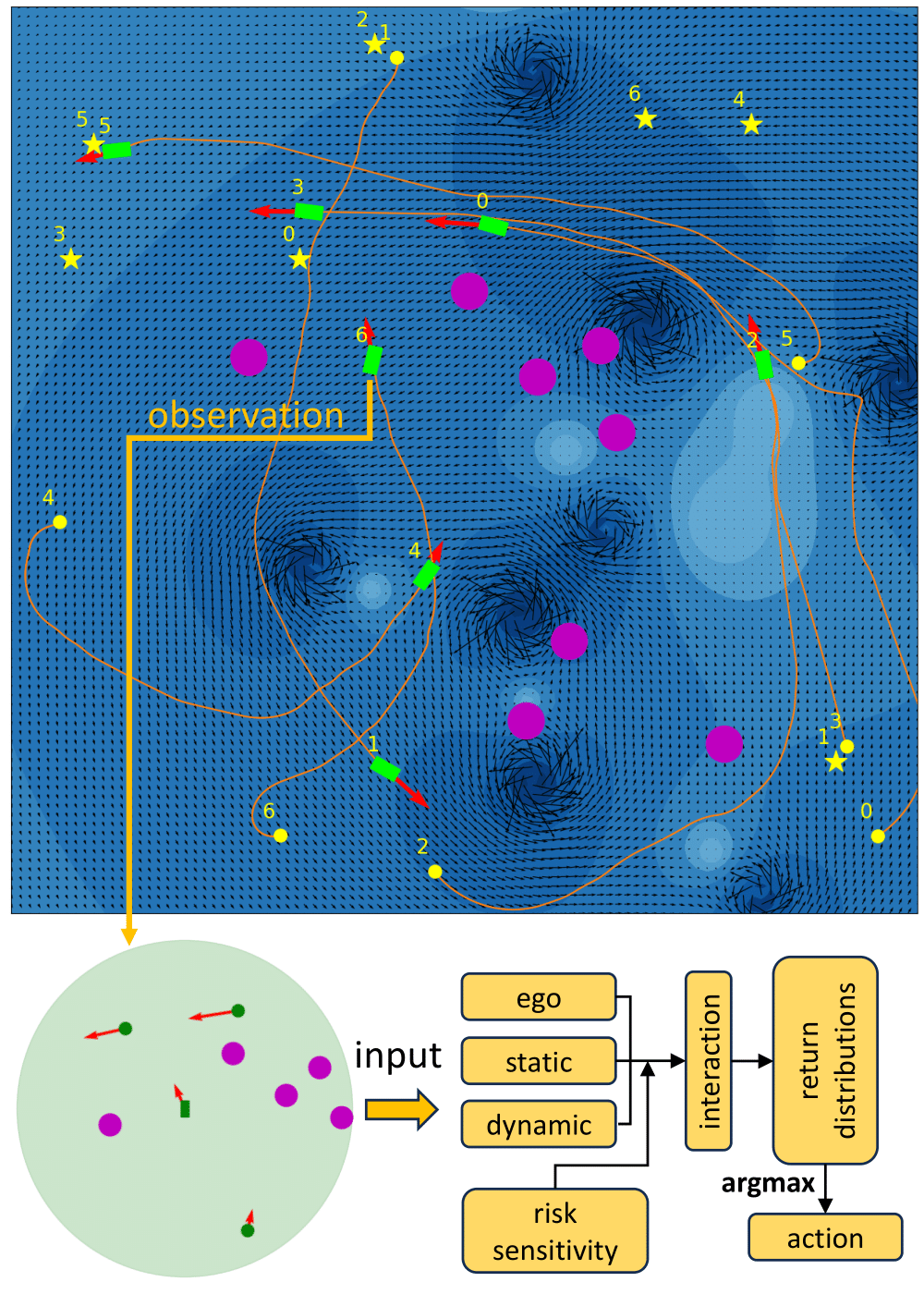}
    \caption{\textbf{Our decentralized decision making framework.}
    }
    \label{fig:demonstration}
    \vspace{-8mm}
\end{figure}

Deep Reinforcement Learning (DRL) combines the ability of Reinforcement Learning to adapt 
to unknown environments through trial-and-error learning \cite{sutton2018reinforcement} with the strong representation capability of deep neural networks, and has been applied to various robotics control problems, including to ASVs \cite{cheng2018concise, woo2020collision}.
Different from DRL that aims to learn the expected return of actions, Distributional Reinforcement Learning (Distributional RL) learns \textit{return distributions} \cite{bellemare2017distributional}, enabling the deployment of risk-sensitive policies by distorting return distributions with risk measures \cite{dabney2018implicit} for enhanced safety performance.

Building on our previous work on single ASV sensor-based navigation with Distributional RL \cite{lin2023usv}, we propose a Distributional RL based policy for decentralized multi-ASV collision avoidance with the following contributions: (1) We design a policy network structure that encodes interaction relations among the ego vehicle, static obstacles and other vehicles, and develop a decentralized multi-agent training framework that induces a coordinated collision avoidance policy by parameter sharing among training agents. 
(2) The proposed policy based on Distributional RL is compared against a traditional DRL policy and classical methods, including Artificial Potential Fields (APF) and Reciprocal Velocity Obstacles (RVO), in simulation experiments. The results indicate that the proposed framework achieves the highest success rate over all levels of environmental complexity, with nearly the smallest amount of travel time and energy cost.
When equipped with a proposed variant of our algorithm that automatically adapts risk sensitivity, the navigation safety of the proposed policy can be clearly improved in highly congested environments.
(3) Our approach has been made freely available at \url{https://github.com/RobustFieldAutonomyLab/Multi_Robot_Distributional_RL_Navigation}.


\vspace{-1mm}
\section{Related Works}
\label{sec:related works}

Traditional non-RL-based methods are widely used to solve the problem of navigating multiple robots in a decentralized manner.
Since the original proposal of using Artificial Potential Fields (APF) in mobile robot navigation \cite{khatib1986real}, it has also been used in multi-robot scenarios, by considering other robots as moving obstacles and generating repulsive forces according to their positions and velocities \cite{sun2017collision, fan2020improved}.
Fiorini et al. \cite{fiorini1998motion} applied the concept of Velocity Obstacles to the dynamic obstacle avoidance problem, which identifies the set of velocities leading to future collisions with moving obstacles. 
Reciprocal Velocity Obstacles (RVO) \cite{van2008reciprocal, guy2009clearpath} and Optimal Reciprocal Collision Avoidance (ORCA) \cite{van2011reciprocal, alonso2013optimal} improved the performance of velocity obstacle methods on the multi-robot navigation problem by selecting velocities that exhibit reciprocal obstacle avoidance behaviors.
Kim et al. \cite{kim2015brvo} proposed to predict the future motion of neighboring agents with past observations of them by Ensemble Kalman Filtering, showing the robustness of the method to noisy, low-resolution sensors, and missing data.  
To solve the freezing robot problem when navigating in cluttered environments, Trautman et al. \cite{trautman2010unfreezing, trautman2013robot} modeled the interactions among agents with Gaussian processes to achieve cooperative navigation.

Decentralized multi-agent collision avoidance solutions based on Deep Reinforcement Learning (DRL) have been developed in recent years.
Chen et al. \cite{chen2017decentralized, chen2017socially} proposed DRL based methods that aim at minimizing the expected time to goal while avoiding collisions, and can show socially aware behaviors in pedestrian-rich environments.
Everett et al. \cite{everett2018motion} deal with varying numbers of observed agents by using long short-term memory
(LSTM) structures, which improves the policy performance when the number of agents increases. 
Long et al. \cite{long2018towards} employed a centralized learning and decentralized execution framework that trains agents capable of navigating in complex environments with a large number of robots.
Chen et al. \cite{chen2019crowd, chen2020relational} modeled the higher-order interactions among agents, which consider not only the direct interactions between the ego agent and other agents, but also the indirect effects of interactions among other agents on the ego agent.
In addition to the observations of other agents, Liu et al. \cite{liu2020robot} included an additional occupancy grid map or an angular map to encode static obstacle information, improving navigation performance in environments cluttered with static obstacles. 
Han et al. \cite{han2022reinforcement} trained DRL agents that exhibit reciprocal collision avoidance behaviors by designing a reward function based on RVO.

Compared to DRL, Distributional RL methods capture \textit{return distributions} instead of only the expected return \cite{bellemare2017distributional}, and these methods have been used to train risk-sensitive agents for a variety of navigation tasks \cite{choi2021risk, kamran2021minimizing, liu2023adaptive, lin2023route, lin2023usv} in recent years. These works focus on the single robot navigation problem in environments with static obstacles or dynamic obstacles moving along pre-defined trajectories. We explore the application of Distributional RL to the \textit{decentralized multi-robot navigation problem} in this paper.

\vspace{-1mm}
\section{Problem Setup}
\label{sec:problem setup}
Multiple robots are required to navigate in a 
simulated marine environment with no prior information describing current flows and static obstacles. 
At each time step $t$, each robot receives an observation of its own state, and of the states of static obstacles and other robots within the detection range, which can be expressed as a combined state vector $\mathbf{s}_t=[\mathbf{s}_{ego}^t,\mathbf{s}_{static}^t.\mathbf{s}_{dynamic}^t]$, where $\mathbf{s}_{ego}$ contains the goal position and ego velocity, $\mathbf{s}_{static}$ contains the position and radius of detected obstacles, and $\mathbf{s}_{dynamic}$ contains the position and velocity of other detected robots. 
\begin{equation}
    \mathbf{s}_{ego} = [p_{x}^{goal},p_{y}^{goal},v_x,v_y]
\end{equation}
\begin{equation}
    \mathbf{s}_{static} = [\mathbf{o}_s^1,\dots,\mathbf{o}_s^m],\ \mathbf{o}_s^i=[p^i_{sx},p^i_{sy},r^i_s]
\end{equation}
\begin{equation}
    \mathbf{s}_{dynamic} = [\mathbf{o}_d^1,\dots,\mathbf{o}_d^n],\ \mathbf{o}_d^i=[p^i_{dx},p^i_{dy},v^i_{dx},v^i_{dy}]
\end{equation}
The pose of each robot can be represented as $(x,y,\theta)$, where $(x,y)$ are the global Cartesian coordinates of the robot, and $\theta$ is its orientation. 
As shown in Fig. \ref{fig:demonstration}, $\mathbf{s}_t$ is expressed in the robot frame.
The objective is to obtain a policy $\pi(\mathbf{s})$ that minimizes the expected travel time to a specified goal, while avoiding collisions with static obstacles and other robots. 
\begin{equation}
\begin{aligned}
    \arg\min_{\pi(\mathbf{s})}\ &\mathbb{E}[t_{goal}|\mathbf{s}_0,\pi], \\
    s.t.\quad &||\mathbf{p}_s^{i,t}||> r_s^{i,t} + r_{rob}\quad \forall\ t,\ \mathbf{o}_s^{i,t}\in\mathbf{s}_{static}^t \\
    &||\mathbf{p}_d^{i,t}||> 2\cdot r_{rob}\quad \forall\ t,\ \mathbf{o}_d^{i,t}\in\mathbf{s}_{dynamic}^t \\
    &||\mathbf{p}^{goal,t_{goal}}||<=d_{thres}
\end{aligned}
\end{equation}
In the above formula, $\mathbf{p}_s=[p_{sx},p_{sy}]$, $\mathbf{p}_d=[p_{dx},p_{dy}]$,$r_{rob}$ is the collision radius of the robot, and the robot is considered to reach its goal if its Euclidean distance to the goal is within the threshold $d_{thres}$.

The simulation environment we use is similar to our prior work \cite{lin2023usv}. 
The Rankine vortex model \cite{acheson1991elementary} is utilized to simulate current flows, where the angular velocity of the vortex core $\Omega=\Gamma / (2\pi r_0^2)$. 
\begin{equation}
    \label{eq:rankine}
    v_r = 0,\quad v_\theta(r) = \frac{\Gamma}{2\pi}\left\{
            \begin{array}{lr}
                 r/r_0^2, &\text{if}\ r\leq r_0 \\
                 1/r, &\text{if}\ r>r_0
            \end{array}
        \right.
\end{equation}

The kinematic model described in \cite{lolla2014time} is used to model the effect of current flow on robot motion.  
\begin{equation}
    \label{eq:pointrobot}
    \frac{d\mathbf{X}(t)}{dt} = \mathbf{V}(t) =  \mathbf{V}_C(\mathbf{X}(t)) +\mathbf{V}_S(t)
\end{equation}
$\mathbf{V}(t)$ is the total velocity of the robot, $\mathbf{V}_C(\mathbf{X}(t))$ is the current flow velocity at robot position $\mathbf{X}(t)$, and $\mathbf{V}_S(t)$ is the robot steering velocity. The robot action at time step $t$ is $\mathbf{a}_t=(a_t,w_t)$, where $a$ is the rate of change in the magnitude of $\mathbf{V}_S(t)$, and $w$ is the rate of change in the direction of $\mathbf{V}_S(t)$.
Then the navigation policy can be expressed as $\pi: \mathbf{s}\rightarrow(a,w)$. 
Linear acceleration $a$ and angular velocity $w$ may be selected as follows: $a \in \{-0.4,0.0,0.4\} \text{ m}/\text{s}^{2}$, and $w \in \{-0.52,0.0.0.52\} \text{ rad}/\text{s}$.
The forward speed is clipped to $[0,v_{max}]$.

\vspace{-1mm}
\section{Approach}
\label{sec:Approach}
\subsection{Reinforcement Learning}
We formulate the problem as a Markov Decision Process $(\mathcal{S},\mathcal{A},P,R,\gamma)$, where $\mathcal{S}$ and $\mathcal{A}$ are the sets of states and actions of the agents described in Section \ref{sec:problem setup}.
$P(s'|s,a)$ is the state transition function reflecting the environment dynamics introduced in Section \ref{sec:problem setup}, and is unknown to the agent due to the unobservable current flow and intents of the other agents. $R(s,a)$ is the reward function, and $\gamma\in [0,1)$ is the discount factor.
At each time step $t$, given the observation of the current state $s_t$, the agent chooses an action $a_t$, which causes the transition to the state $s_{t+1}\sim P(\cdot|s_t,a_t)$ and receiving the reward $r_{t+1}=R(s_{t+1},a_{t+1})$.  
The action value function $Q^\pi(s, a)$ is defined as the expected return of taking action $a$ at state $s$ and following the policy $\pi$ thereafter.
\begin{equation}
  \label{eq:expected return}
  Q^\pi(s, a) = \mathbb{E}_\pi[\sum_{k=0}^\infty \gamma^k r_{t+k+1}| s_t=s, a_t=a]
\end{equation}
An optimal policy $\pi_*$ maximizes $Q^\pi(s,a)$, and the resulting optimal action value function $Q^{\pi_*}(s, a)$ satisfies the Bellman optimality equation \eqref{eqn:Bellman optimality equation}.
\begin{equation}
    \label{eqn:Bellman optimality equation}
        Q^{\pi_*}(s, a) = \mathbb{E}[r_{t+1}+\gamma\max_{a'}Q^{\pi_*}(s', a')]
\end{equation}
The reward function we use is shown as follows, where $r_{step}=-1.0$ and $r_{foward,t}=||\mathbf{p}^{goal,t-1}-\mathbf{p}^{goal,t}||$ encourage the agent to move towards the goal, $\mathbf{1}$ is the indicator function, and $r_{collision}=-50.0$ or $r_{goal}=100.0$ is given when the agent reaches the goal or collides, respectively. 
\begin{equation}
\begin{array}{cl}    
r_t=&r_{step}+r_{foward,t}+\mathbf{1}_{collision}(s_t)\cdot r_{collision} \\
&+\mathbf{1}_{reach\_goal}(s_t)\cdot r_{goal}
\end{array}
\end{equation}

\subsection{Traditional DRL}
DQN \cite{mnih2015human} represents the action value function as a neural network model $Q(s,a;\theta)$, and learns the network parameters $\theta$ by optimizing the loss $\mathcal{L}_{\text{DQN}}$ computed with samples in the format of $(s,a,r,s')$ from experiences.
\begin{equation}
    \label{eq:DQN loss}
    \mathcal{L}_{\text{DQN}} = \mathbb{E}[(r+\gamma\max_{a'}Q(s',a';\theta^-)-Q(s,a;\theta))^2]
\end{equation}
\subsection{Distributional RL}
Instead of the expected return $Q^\pi(s,a)$, Distributional RL algorithms \cite{bellemare2017distributional} learn the return distribution $Z^\pi(s,a)$, where $Q^\pi(s,a) = \mathbb{E}[Z^\pi(s,a)]$, and the distributional Bellman equation is considered.
\begin{equation}    
\label{eqn:Distributional bellman}
    Z^\pi(s,a) \overset{D}{=} R(s,a) + \gamma Z^\pi(s',a') 
\end{equation}

Our planner is based on implicit quantile networks (IQN) \cite{dabney2018implicit}, which expresses the return distribution with a quantile function $Z_\tau := F_Z^{-1}(\tau)$, where $\tau\sim U([0,1])$, and enables the incorporation of a distortion risk measure $\beta:[0,1]\rightarrow[0,1]$ to compute a risk distorted expectation $Q_\beta(s,a)$ as well as a risk-sensitive policy $\pi_\beta$. We use $K=32$.
\begin{equation}
    Q_\beta(s,a) = \mathbb{E}_{\tau\sim U([0,1])}[Z_{\beta(\tau)}(s,a)]
\end{equation}
\begin{equation}
    \label{eq:approximate pi beta}
    \pi_\beta(s) = \text{argmax}_a \frac{1}{K}\sum_{k=1}^K Z_{\beta(\Tilde{\tau}_k)}(s,a),\ \Tilde{\tau}_k\sim U([0,1])
\end{equation}
Network parameters of the IQN model can be learned by optimizing the loss $\mathcal{L}_{\text{IQN}}$. We use $N=N'=8$, $\kappa=1.0$. 
\begin{equation}
    \label{eq:sampled TD}
    \delta^{\tau_i,\tau'_j} = r + \gamma Z_{\tau'}(s',\pi_\beta(s')) - Z_{\tau}(s,a)
\end{equation}
\begin{equation}
    \label{eq:quantile Huber}
    \begin{array}{c}
        \rho_\tau^\kappa(u) = |\tau-\mathbf{1}_{\{u<0\}}|(\mathcal{L}_\kappa(u)/\kappa), \\[5pt]

        \text{where }\mathcal{L}_\kappa(u) = \left\{
            \begin{array}{lr}
                 \frac{1}{2}u^2, &\text{if}\ |u|\leq\kappa \\
                 \kappa(|u|-\frac{1}{2}\kappa), & \text{otherwise} 
            \end{array}
        \right.
    \end{array}
\end{equation}
\begin{equation}
    \label{eq:IQN loss}
    \mathcal{L}_{\text{IQN}} = \frac{1}{N'}\sum_{i=1}^N \sum_{j=1}^{N'} \rho_{\tau_i}^\kappa(\delta^{\tau_i,\tau'_j})
\end{equation}

\subsection{Training Decentralized Policy in Multi-Agent Scenarios}

\begin{figure}
    \centering
    \includegraphics[width=\linewidth]{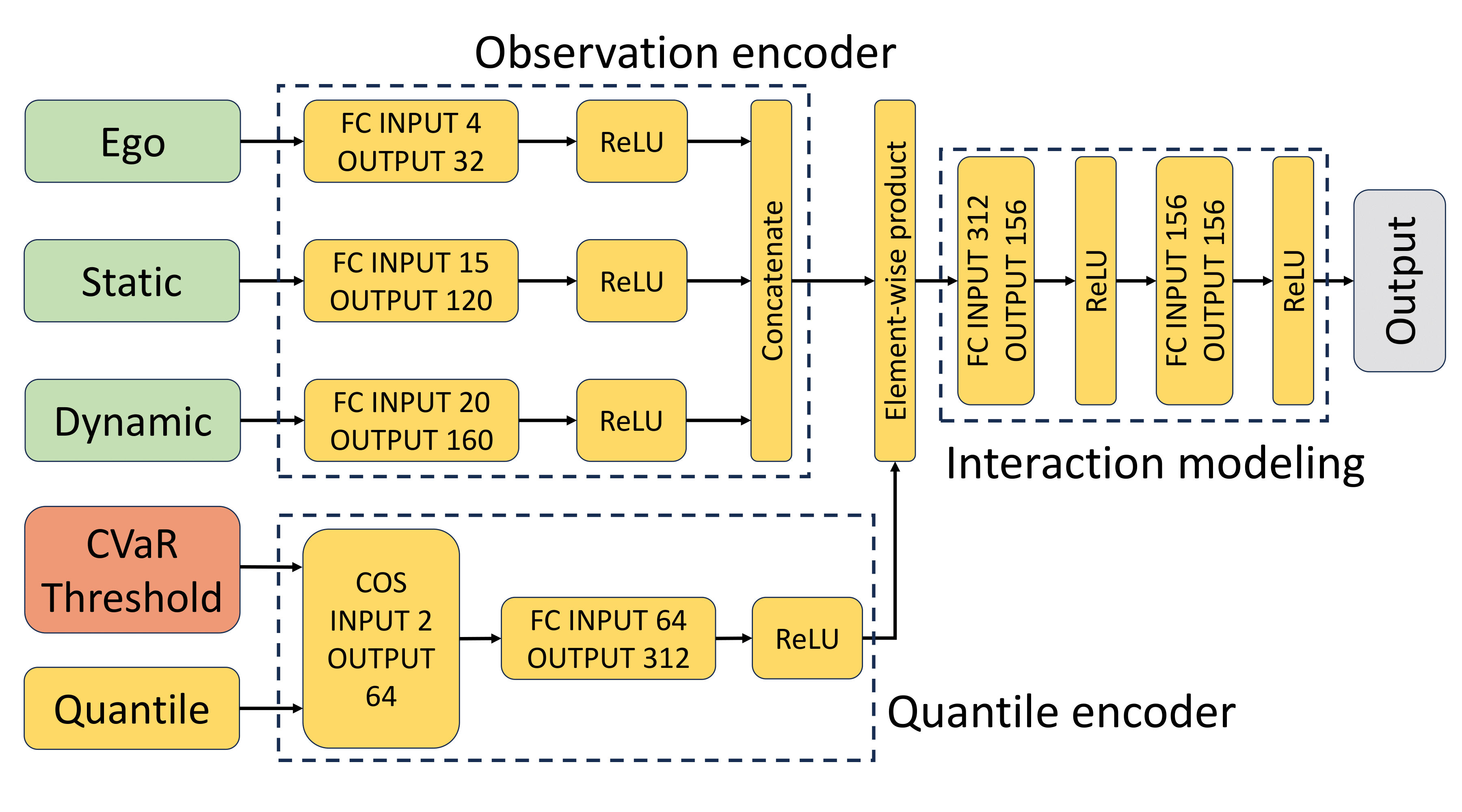}
    \caption{\textbf{IQN network model.} FC, COS and ReLU stand for fully connected layer, cosine embedding layer, and rectified linear unit.
    Outputs of the model are the return distributions of actions. 
    }
    \label{fig:network}
    \vspace{-4mm}
\end{figure}

The network architecture we use for the IQN agent is shown in Fig. \ref{fig:network}.
To create a fixed-size input for the model, the observed static obstacles and other robots are sorted according to the distance to the ego robot, and maximum $m=5$ static obstacles and $n=5$ other robots closest to the ego robot are considered in the input.
The input is padded with zeros if the number of observed static obstacles and other robots are smaller than $m$ and $n$.
Observations of the ego robot, static obstacles and other robots are transformed into corresponding latent features through the observation encoder.  
To encode risk sensitivity, the Conditional Value at Risk (CVaR) \eqref{eq:cvar} is used, which scales the quantile sample $\tau$ with the CVaR threshold $\phi$, and uses it to compute a set of cosine features \eqref{eq:cos}.
After observation features are combined with risk sensitivity features, they are passed through two fully connected layers to model interactions among the ego robot, static obstacles and other robots, and outputs the return distributions of actions.    
The DQN agent uses the same network model as Fig. \ref{fig:network}, except that risk sensitivity and the quantile encoder are not used, and its output is composed of action values.
\begin{equation}
    \label{eq:cvar}
    f(\tau;\phi) = \phi\tau,\quad \phi\in(0,1],\ \tau\sim U([0,1])    
\end{equation}
\begin{equation}
    \label{eq:cos}
    [\cos(\pi\cdot 0\cdot f(\tau;\phi)),\dots, \cos(\pi\cdot 63\cdot f(\tau;\phi))]
\end{equation}

\begin{algorithm}
$\mathcal{R}, \{s_i|Rob_i\in\mathcal{R}\}\leftarrow$ initialize env, $l_{episode} \leftarrow 0$ \\ 
$\mathcal{M}\leftarrow\emptyset,\mathcal{\pi}\leftarrow\text{DQN or IQN}, \mathcal{L}\leftarrow\mathcal{L}_{\text{DQN}}\text{ or }\mathcal{L}_{\text{IQN}}$ \\
\For{$t=1,\dots,t_{total}$}{
    $\epsilon\leftarrow f_{linear}(t,t_{total},\epsilon_{max},\epsilon_{min})$ \\
    $\mathcal{A}\leftarrow \{a_i=f_{\epsilon-greedy}(\pi_\theta(s_i),\epsilon)|Rob_i\in\mathcal{R}\}$ \\
    $\{(s'_i,r_i)|Rob_i\in\mathcal{R}\}\leftarrow \text{execute }\mathcal{A}\text{ in env}$ \\
    $\mathcal{M}\leftarrow$ add $\{(s_i,a_i,r_i,s'_i)|Rob_i\in\mathcal{R}\}$ \\
    $\mathcal{R}\leftarrow\mathcal{R}\setminus\{Rob_i|Rob_i\text{ reaches goal or collides}\}$ \\
    \If{$t\bmod t_{learn\_freq}=0$}{
         $\mathcal{B}=\{(s,a,r,s')_k|k=1,\dots n_{batch}\}\leftarrow$ sample batch from $\mathcal{M}$\\
         $\theta\leftarrow$ optimize $\mathcal{L}(\mathcal{B},\pi_\theta)$ w.r.t $\theta$ 
    }
    \If{$t\bmod t_{eval\_freq}=0$}{
        evaluate $\pi_\theta$
    }
    \If{$l_{episode} > l_{episode\_max}\ or\ \mathcal{R}=\emptyset$}{
        $\mathcal{R}, \{s_i|Rob_i\in\mathcal{R}\}\leftarrow$ reset env,
        $l_{episode} \leftarrow 0$
    }
    \Else{
        $s_i\leftarrow s'_i$ for all $Rob_i\in\mathcal{R}$, $l_{episode}\mathrel{+}=1$
    }
}

\caption{Policy training algorithm}
\label{Alg:training}
\end{algorithm}

\begin{table}
    \centering
    \caption{\textbf{Training schedule.} Hyperparameters used to create training environments.}
    \begin{tabular}{ccccccc}
         \toprule
         Timesteps (million) & 1st & 2nd & 3rd & 4st & 5st & 6st  \\ \midrule
        Number of robots & 3 & 5 & 7 & 7 & 7 & 7 \\ 
        Number of vortices & 4 & 6 & 8 & 8 & 8 & 8 \\ 
        Number of obstacles & 0 & 0 & 2 & 4 & 6 & 8 \\ 
        Min distance between & \multirow{2}{*}{30.0} & \multirow{2}{*}{35.0} & \multirow{2}{*}{40.0} & \multirow{2}{*}{40.0} & \multirow{2}{*}{40.0} & \multirow{2}{*}{40.0} \\
        start and goal & & & & & & \\ \bottomrule
    \end{tabular}
    \label{tab:schedule}
    \vspace{-3mm}
\end{table}

\begin{figure}
    \centering
    \includegraphics[width=\linewidth]{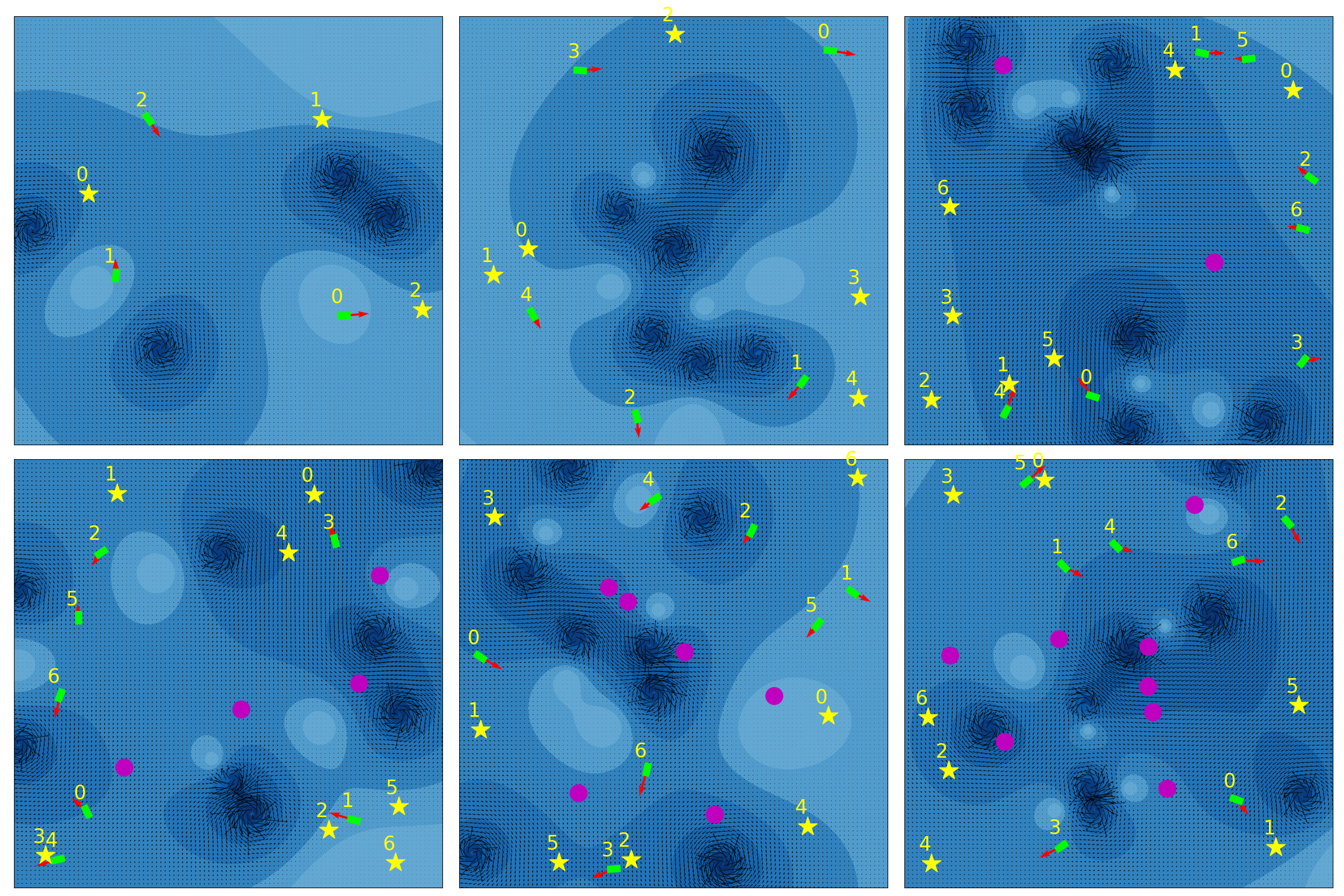}
    \caption{\textbf{Training environments.} 
    Examples of random environments generated according to the training schedule (environments of increasing difficulty are shown from left to right, top to bottom). 
    The initial poses and velocities of the robots are indicated with green rectangles and red arrows. 
    }
    \vspace{-3mm}
    \label{fig:envs}
\end{figure}

\begin{figure}[!]
    \centering
    \includegraphics[width=\linewidth]{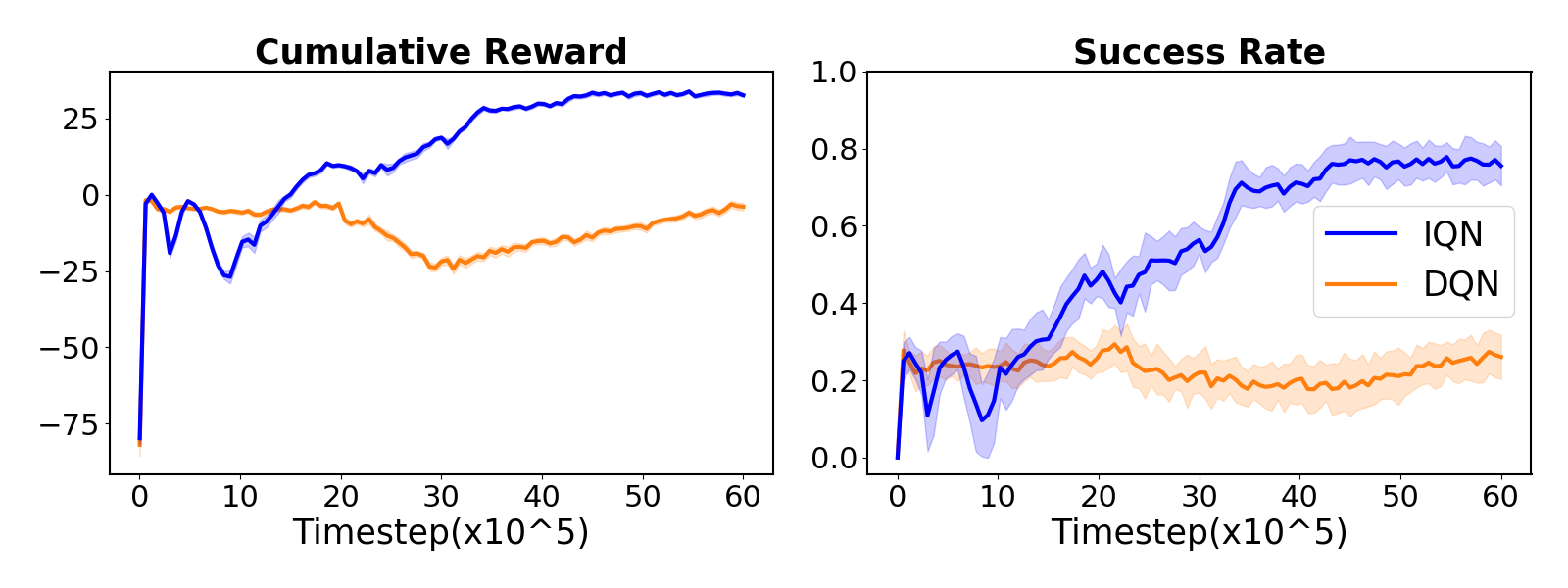}
    \caption{\textbf{Evaluation performance during training.} Solid lines and bandwidths indicate the mean and standard error over the results of all learning models.
    }
    \label{fig:learning curves}
    \vspace{-6mm}
\end{figure}

The process of training IQN and DQN agents is described in Algorithm \ref{Alg:training}.
At the beginning of every training episode, a new training environment with randomly generated robots, vortices and static obstacles is created according to the training schedule shown in Table \ref{tab:schedule}, which is designed to gradually increase the complexity of the environment as training proceeds.
Example training environments with different levels of complexity are shown in Fig. \ref{fig:envs}.
The radius of each static obstacle is set to 1.0 meters. 

To induce coordinated obstacle avoidance behavior among robots, we maintain only one learning model, which is shared with all robots during their individual decision making processes.
We use $\phi=1.0$ for the IQN model during training.
An $\epsilon$-greedy policy is employed in training, where the exploration rate $\epsilon$ decays linearly from $\epsilon_{max}=0.6$ to $\epsilon_{min}=0.05$ in the first 25\% of training time steps, and remains the same thereafter.
In each step, the experience tuples $(s,a,r,s')$ of all robots are inserted in to the replay buffer $\mathcal{M}$, which is then used to give a batch of samples $\mathcal{B}$ for model training.   
If a robot reaches its goal or collides with static obstacles or other robots, it will be removed from the environment.
The current episode ends when it is longer than $l_{episode\_max}=1000$ or no robot exists in the environment.

For every $t_{eval\_freq}=60,\hspace{-0.5mm}000$ steps, the learning model is evaluated in a set of predefined evaluation environments to understand its learning performance.
We create ten random environments for each level of complexity in Table \ref{tab:schedule}, resulting in a total of sixty evaluation environments.   
We train thirty IQN models and thirty DQN models with different random seeds on an Nvidia RTX 3090 GPU, and show their general performance in Fig. \ref{fig:learning curves}.
If not all robots reach goals at the end, the evaluation episode is considered failed, hence IQN achieves a significantly higher level of safety even in highly congested environments.  

\vspace{-1mm}
\section{Experiments}
\label{sec:experiments}

\begin{figure*}
    \centering
    \begin{subfigure}{0.32\textwidth}
        \centering
        \includegraphics[width=\linewidth]{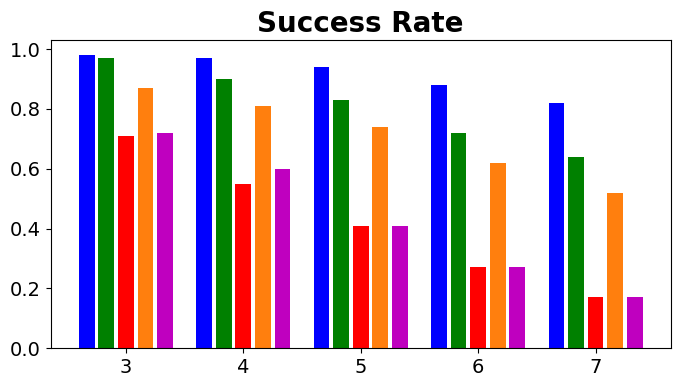}
    \end{subfigure}
    \begin{subfigure}{0.324\textwidth}
        \centering
        \includegraphics[width=\linewidth]{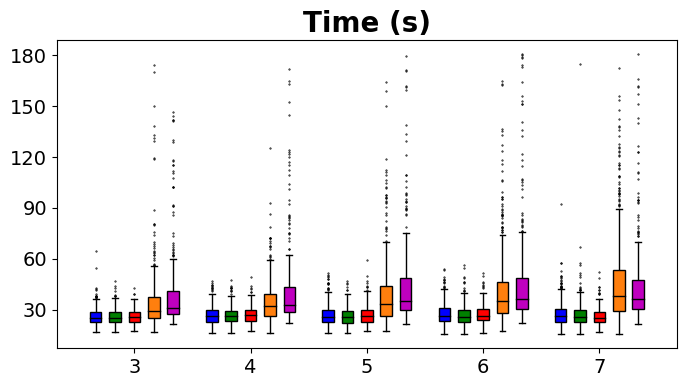}
    \end{subfigure}
    \begin{subfigure}{0.324\textwidth}
        \centering
        \includegraphics[width=\linewidth]{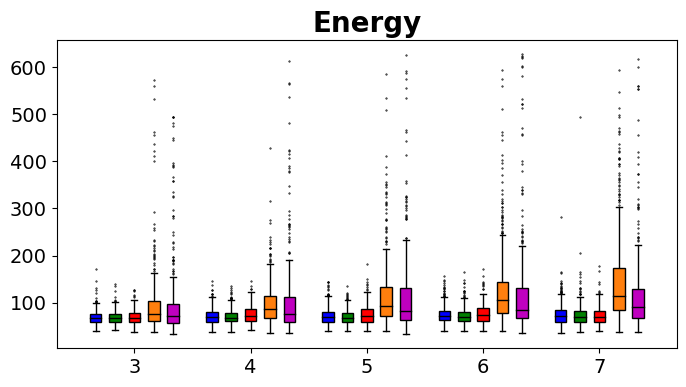}
    \end{subfigure}

    \begin{subfigure}{0.32\textwidth}
        \centering
        \includegraphics[width=\linewidth]{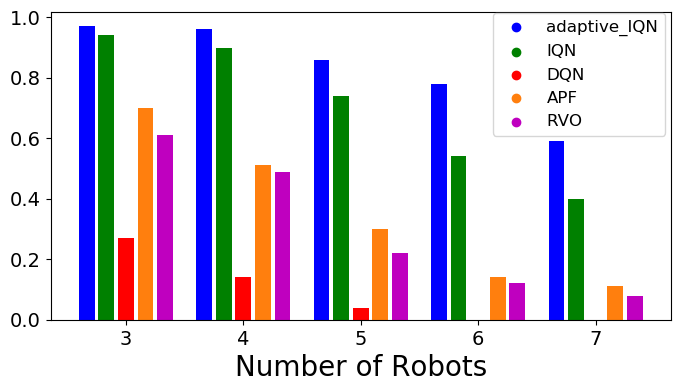}
    \end{subfigure}
    \begin{subfigure}{0.323\textwidth}
        \centering
        \includegraphics[width=\linewidth]{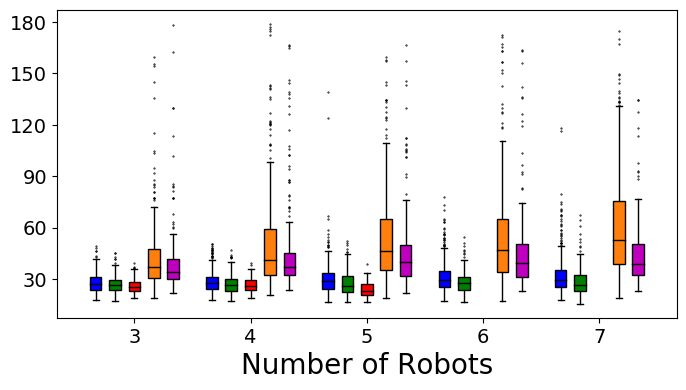}
    \end{subfigure}
    \begin{subfigure}{0.323\textwidth}
        \centering
        \includegraphics[width=\linewidth]{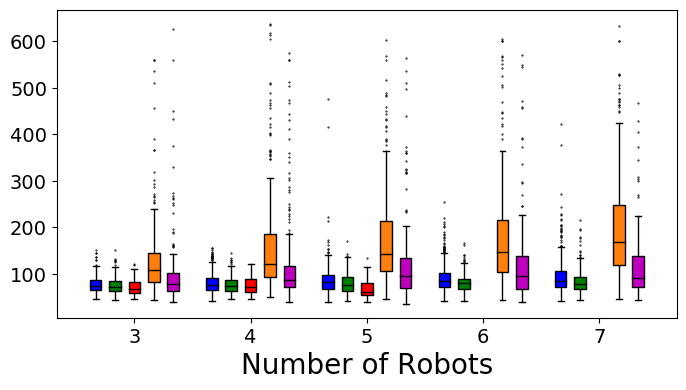}
    \end{subfigure}
    \caption{\textbf{Experimental results.} 
    Top and bottom rows show the results of experiments without and with static obstacles, respectively.
    Time and energy plots show the distributions of travel time and energy consumption of the robots in successful episodes.}
    \label{fig:exp}
    \vspace{-6mm}
\end{figure*}

We evaluate the performance of two randomly selected IQN and DQN models, as well as baseline approaches which will be introduced as follows, on two sets of simulation experiments.
The first set of experiments purely focuses on dynamic obstacle avoidance and does not contain any static obstacles.
500 randomly generated environments of different levels of difficulty (100 per level) are used. The number of robots range from 3 to 7, and the number of vortices range from 4 to 8.
The start and goal of each robot are randomly generated such that the distance is at least 40.0 meters.
In the second set of experiments, static obstacles are also included when creating 500 evaluation environments, where the number of static obstacles ranges from 4 to 8.
If not all robots successfully reach their goals within 3 minutes, the test episode is considered failed. 
The energy consumption of a robot in an experiment episode is computed by summing up the magnitude of all actions executed during the process.

In addition to the greedy policy, we also test the \textit{risk sensitive policy} used in our prior work \cite{lin2023usv}, which adapts the CVaR threshold $\phi$ according to the distance to obstacles.  
\begin{equation}
    \label{eq:adaptive cvar}
    \phi = \left\{
            \begin{array}{lr}
                 \min(d(X,X_O))/d_0, &\text{if}\ \min(d(X,X_O))\leq d_0 \\
                 1.0, &\text{if}\ \min(d(X,X_O))> d_0
            \end{array}
            \right.
\end{equation}
$X$ is the position of the ego robot, and $X_O$ are the positions of all obstacles and other robots.

In addition to IQN and DQN, we also evaluate the performance of two classical methods. 
Fan et al. \cite{fan2020improved} designed an improved Artificial Potential Field (APF) method to deal with both static and dynamic obstacles.
The attractive potential field $U_{\text{att}}$ creates a force that drives the ego robot to the goal.   
\begin{equation}
    U_{\text{att}}(X) = \frac{1}{2} k_{\text{att}} \cdot d^2(X,X_g) 
\end{equation}
The repulsive potential field $U_{\text{rep1}}$ creates a repulsive force to avoid colliding with static obstacles.
\begin{equation}
    \begin{array}{c}
    U_{\text{rep1}}(X) = \left\{
            \begin{array}{lr}
                 U_o(X), &\text{if}\ d(X,X_o)\leq d_0 \\
                 0, &\text{if}\ d(X,X_o)> d_0
            \end{array}
    \right. \\[10pt]
    \text{where } U_o(X) = \frac{1}{2}k_{\text{rep}}(\frac{1}{d(X,X_o)}-\frac{1}{d_0})^2d^n(X,X_g)
    \end{array}
\end{equation}
For dynamic obstacles, an additional repulsive field based on the relative velocity to the ego robot is used. 
\begin{equation}
    \begin{array}{c}
    U_{\text{rep2}}(X,V) = \left\{
            \begin{array}{lr}
                 U_r(X)+U_r(V), &\text{if}\ d(X,X_r)\leq d_0 \\
                 0, &\text{if}\ d(X,X_r)> d_0
            \end{array}
    \right. \\[10pt]
    \text{where } U_r(X) = \frac{1}{2}k_{\text{rep}}(\frac{1}{d(X,X_r)}-\frac{1}{d_0})^2d^n(X,X_g), \\
    U_r(V) = k_v \frac{1}{d(X,X_r)} v_{ao}
    \end{array}
\end{equation}
In the above equations, $X$, $X_g$, $X_o$, and $X_r$ are positions of the robot, goal, static and dynamic obstacles respectively, and $v_{ao}$ is the
relative velocity component in the direction from the ego robot to the dynamic obstacle. 
We use $k_{\text{att}}=50.0$, $k_{\text{rep}}=500.0$, $k_{\text{v}}=1.0$, and $n=2$.
The total force $F = -\nabla U_{\text{att}}(X)-\nabla U_{\text{rep1}}(X)-\nabla U_{\text{rep2}}(X,V)$.
As in our implementation of APFs in previous work \cite{lin2023usv}, we choose the angular velocity that results in the closest alignment with vector $F$ in a single control step, and we choose the linear acceleration closest to the scaled component of $F$ in the direction of robot velocity, as the control action.

Van den Berg et al. \cite{van2008reciprocal} proposed Reciprocal Velocity Obstacles (RVO).
For disc-shaped robots $A$ and $B$, the velocity obstacle $VO^A_B(\mathbf{v}_B)$ is the set of velocities of $A$, $\mathbf{v}_A$, that will lead to collision with $B$ moving at velocity $\mathbf{v}_B$.
Denote a ray starting at $\mathbf{p}$ and pointing in the direction of $\mathbf{v}$ as $\lambda(\mathbf{p},\mathbf{v})=\{\mathbf{p}+t\mathbf{v}|t\geq0\}$, then $VO^A_B(\mathbf{v}_B)$ can be expressed in Equation \eqref{eq:VO}, where $\oplus$ is the Minkowski sum, and -$A$ refers to $A$
reflected about its reference point.
\begin{equation}
\label{eq:VO}
VO_{B}^{A}({{\mathbf{v}}_{B}}) = \lbrace {{\mathbf{v}}_{A}}\left| {\lambda ({{\mathbf{p}}_{A}},{{\mathbf{v}}_{A}} - {{\mathbf{v}}_{B}}) \cap B \oplus - A \ne \emptyset } \right.\rbrace
\end{equation}

\begin{figure*}
    \centering
    \begin{subfigure}{0.49\textwidth}
        \centering
        \includegraphics[width=\linewidth]{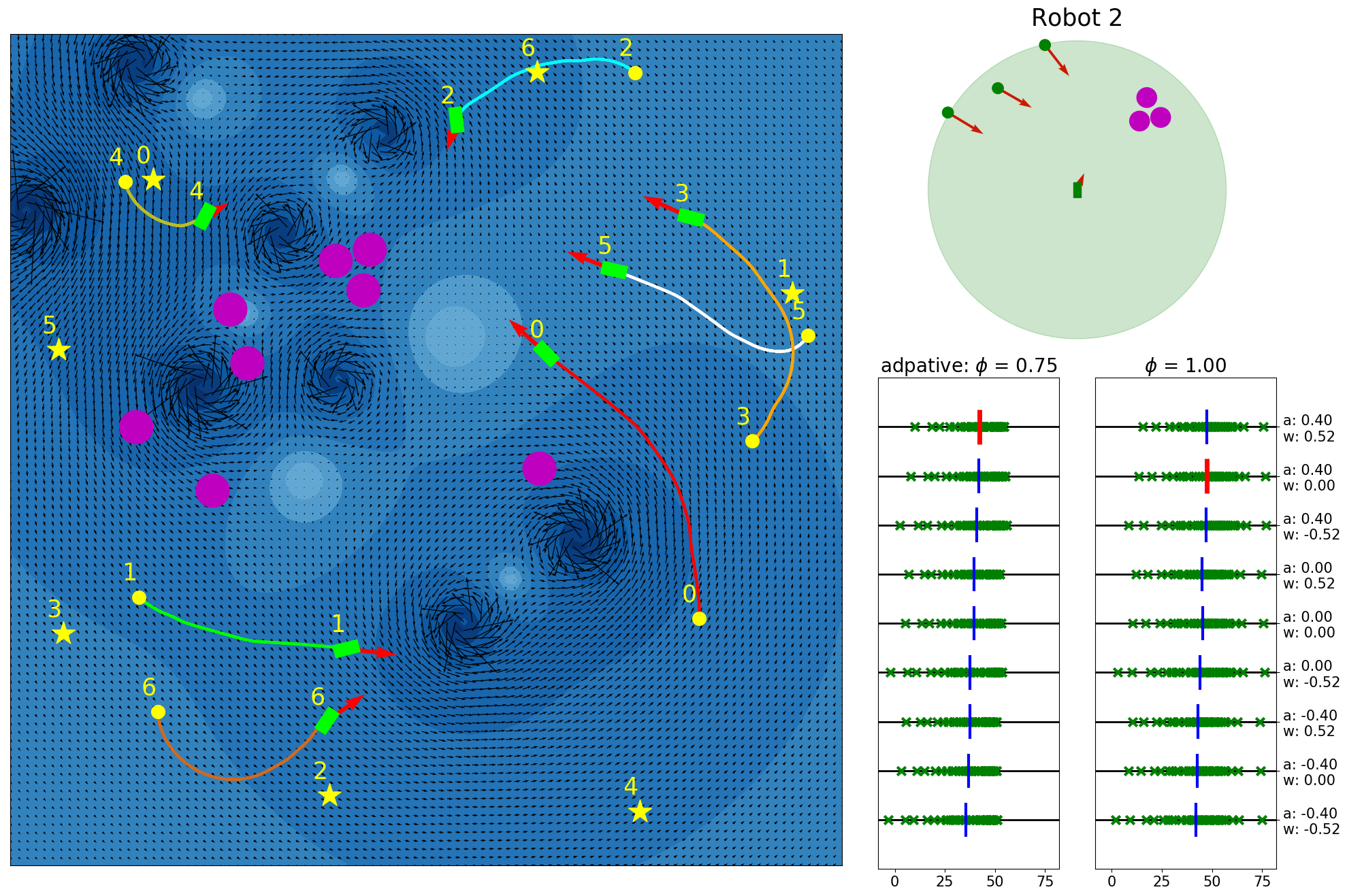}
    \end{subfigure}
    \begin{subfigure}{0.49\textwidth}
        \centering
        \includegraphics[width=\linewidth]{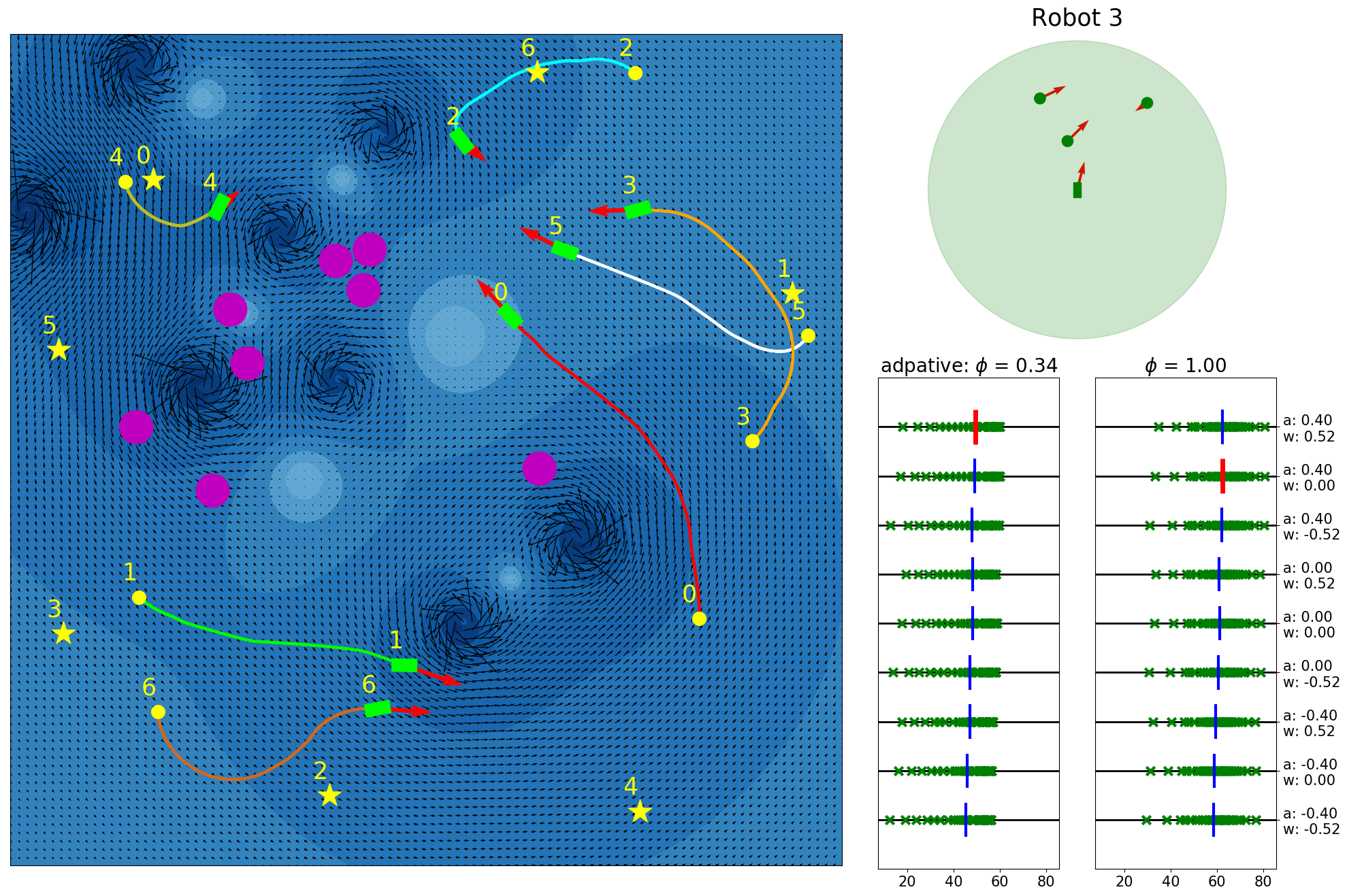}
    \end{subfigure}
    \caption{\textbf{Performance of adaptive IQN agents in congested scenarios.}
    The resulting distributions of adaptive IQN and greedy IQN, in two representative decision-making instances from the same mission, are shown as return values corresponding to sampled quantiles. The mean of each distribution is marked as a vertical bar, and the selected actions are colored in red.
    }
    \label{fig:distributions}
    \vspace{-4mm}
\end{figure*}

\begin{figure*}
    \centering
    \begin{subfigure}{0.195\textwidth}
        \centering
        \includegraphics[width=\linewidth]{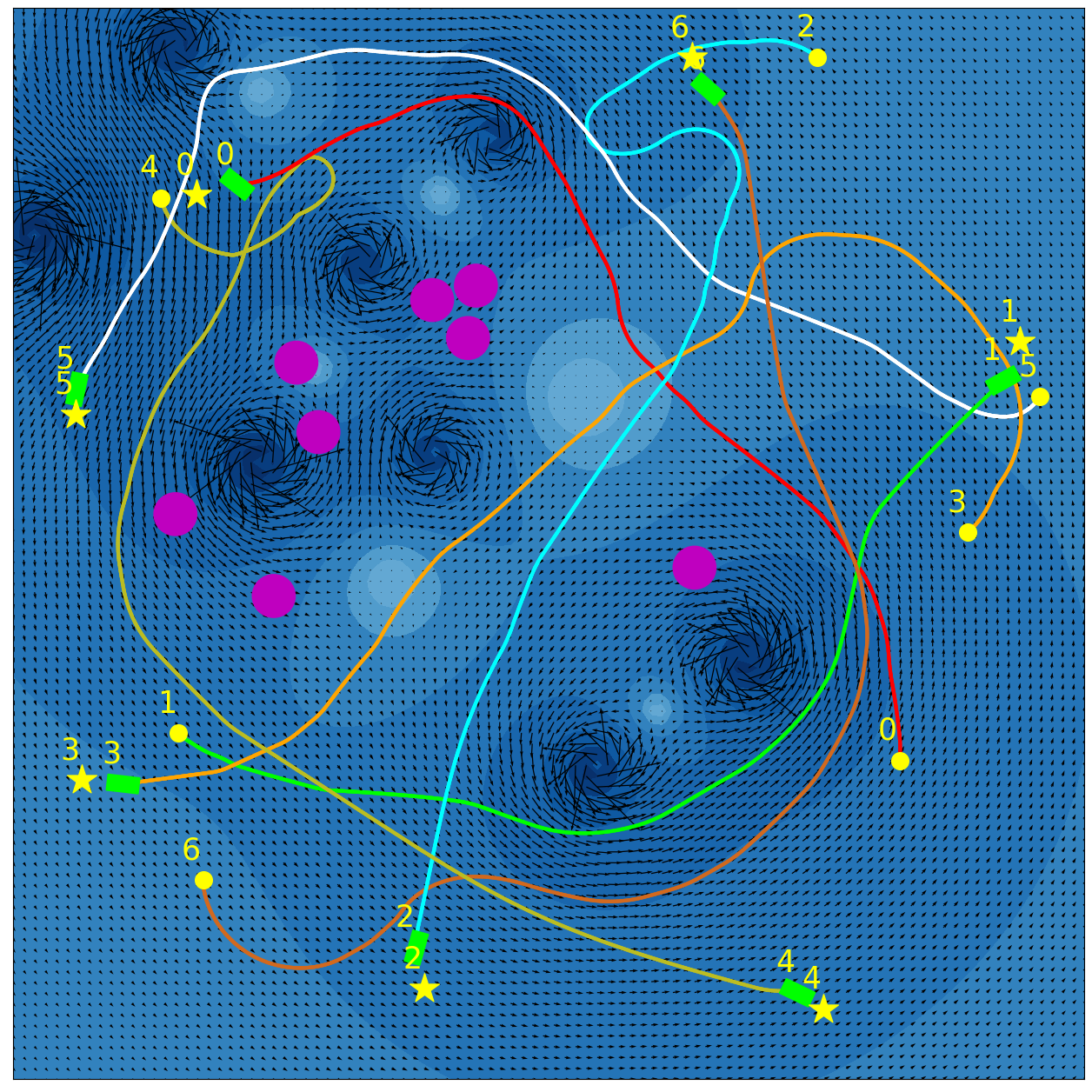}
    \end{subfigure}
    \begin{subfigure}{0.195\textwidth}
        \centering
        \includegraphics[width=\linewidth]{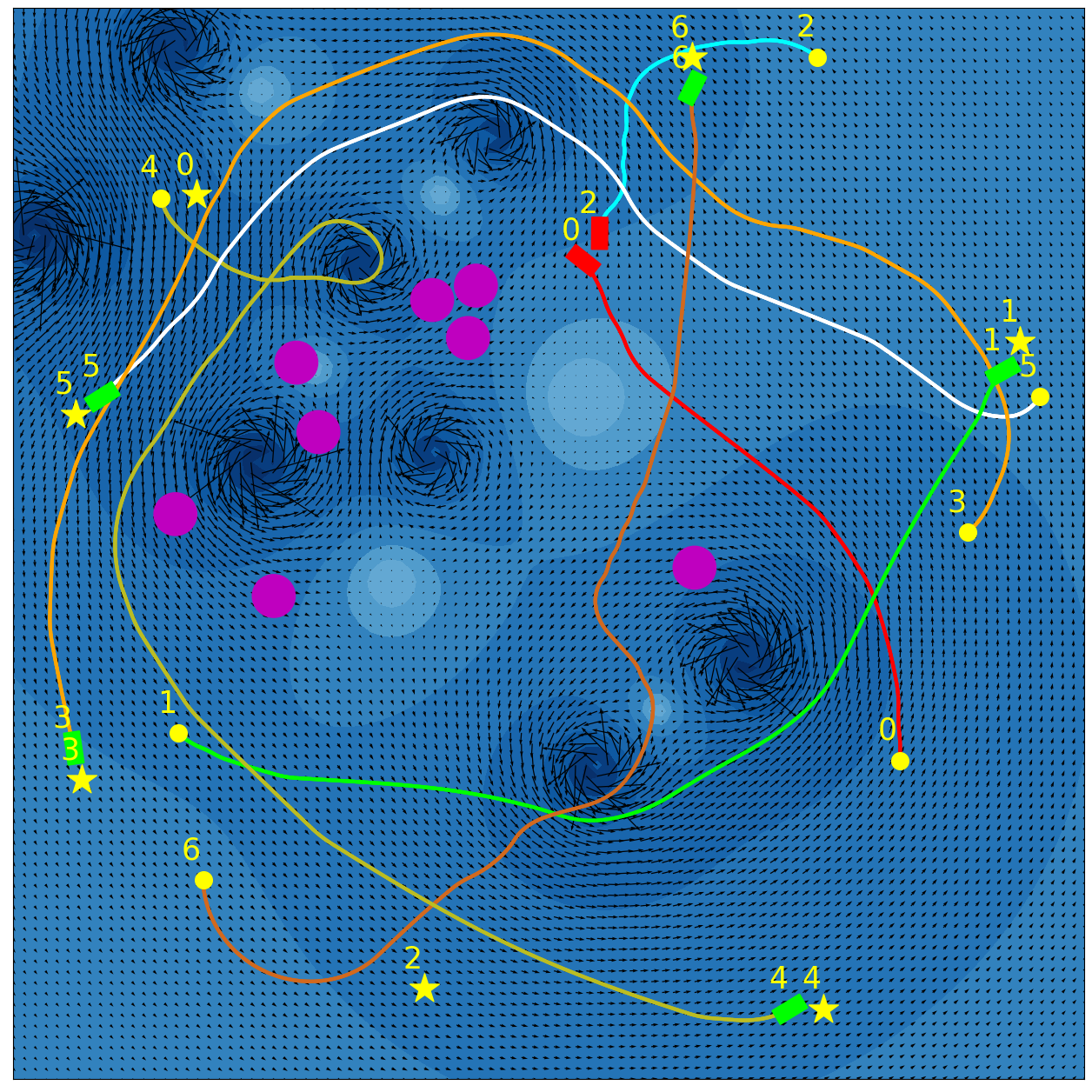}
    \end{subfigure}
    \begin{subfigure}{0.195\textwidth}
        \centering
        \includegraphics[width=\linewidth]{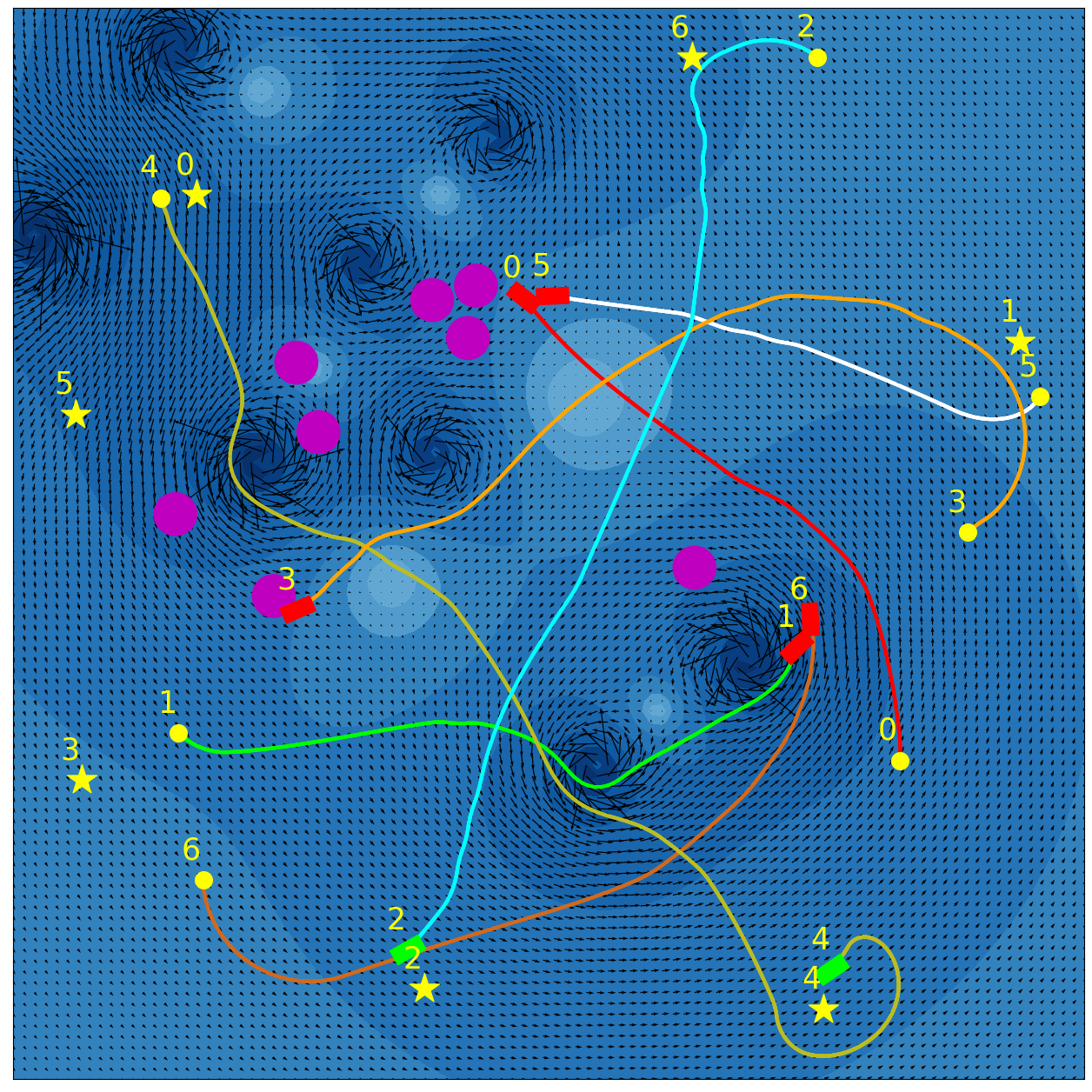}
    \end{subfigure}
    \begin{subfigure}{0.195\textwidth}
        \centering
        \includegraphics[width=\linewidth]{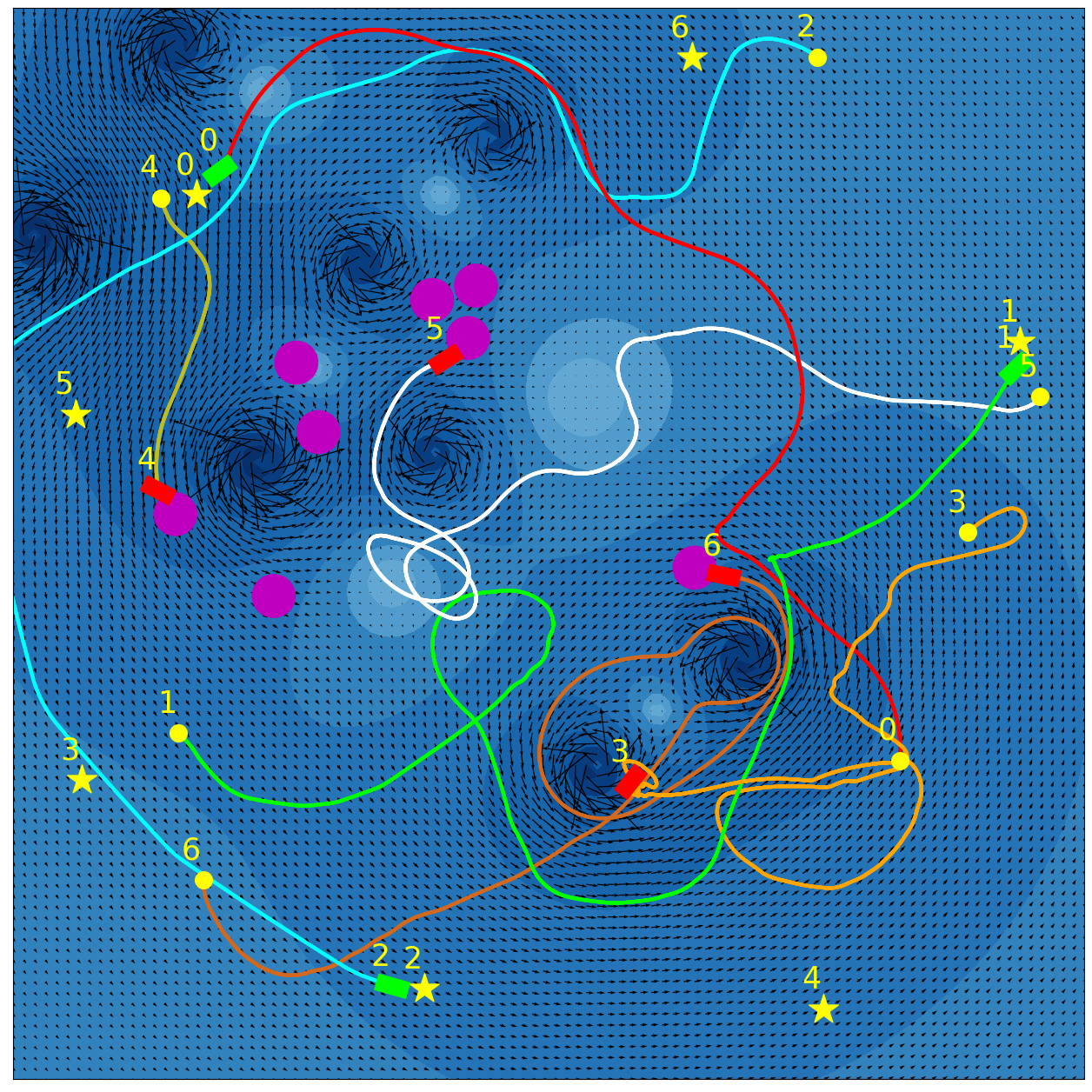}
    \end{subfigure}
    \begin{subfigure}{0.195\textwidth}
        \centering
        \includegraphics[width=\linewidth]{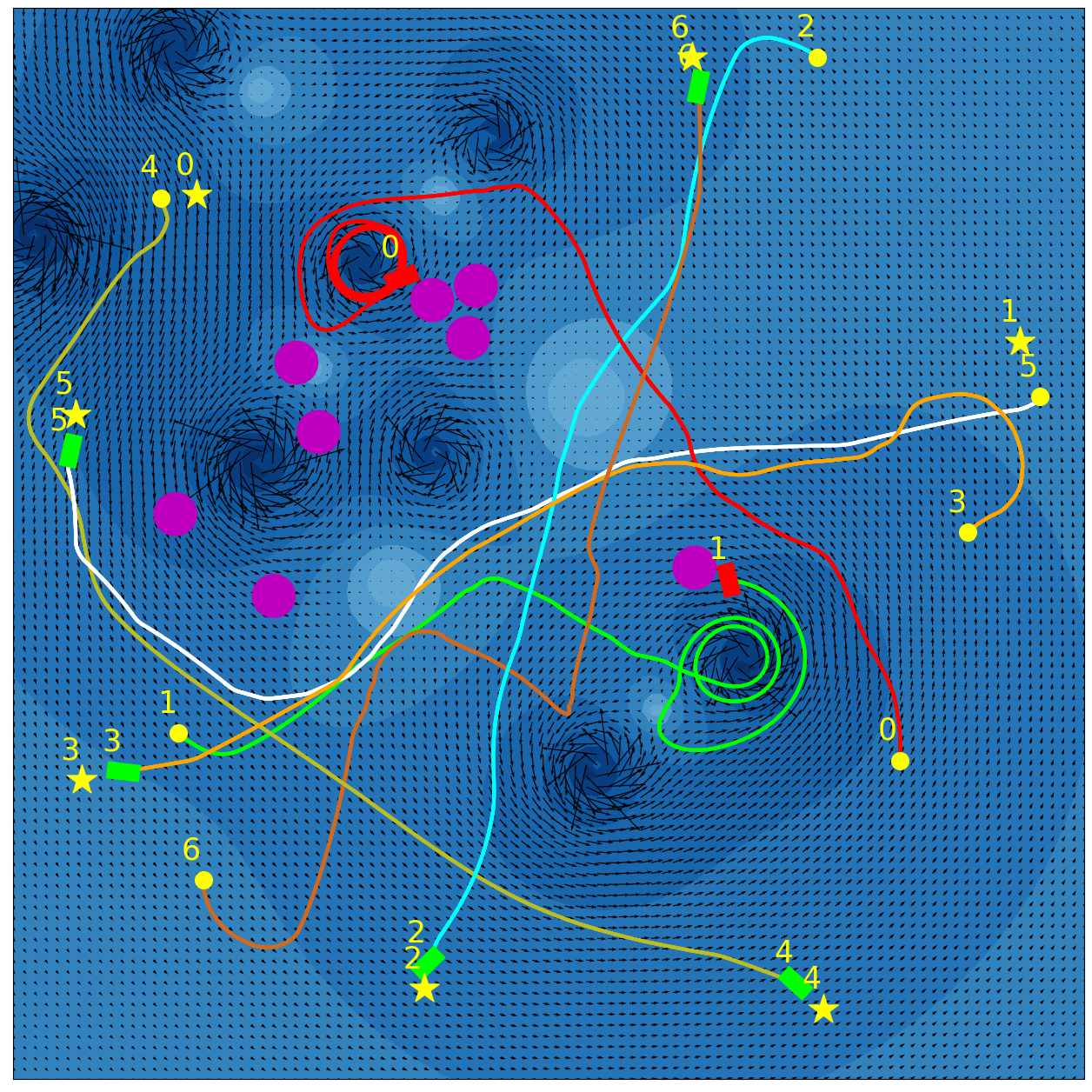}
    \end{subfigure}
    \caption{\textbf{Planned trajectories.}
    Trajectories of adaptive IQN, greedy IQN, DQN, APF, and RVO are shown from left to right, for a representative set of start/goal locations. 
    We mark robots that reach their goals in green, and those that failed in red. 
    }
    \label{fig:trajectories}
    \vspace{-6mm}
\end{figure*}

To solve the problem of oscillatory motions when choosing the velocity that avoids the $VO$s created by other robots, $RVO$ is introduced, which translates the apex of $VO^A_B(\mathbf{v}_B)$ to $({\mathbf{v}}_{A}+{\mathbf{v}}_{B})/2$.
Since no passively moving obstacles exist in our problem, the combined reciprocal velocity obstacle for each robot is the union of all the $RVO$s generated by other robots and the $VO$s generated by static obstacles.
\begin{equation} 
RVO_{B}^{A}({{\mathbf{v}}_{B}},{{\mathbf{v}}_{A}}) = \lbrace {{{{\bf v^{\prime }}}}_{A}}\left| {2{{{{\bf v^{\prime }}}}_{A}} - {{\mathbf{v}}_{A}} \in VO_{B}^{A}({{\mathbf{v}}_{B}})} \right.\rbrace. 
\end{equation}
The velocity $\mathbf{v}'_i$ that minimizes the penalty function \eqref{eq:penalty} is chosen to be the next velocity of robot $i$.  
\begin{equation}
\label{eq:penalty}
penalty_i(\mathbf{v}'_i)=w_i\frac{1}{tc_i(\mathbf{v}'_i)}+||\mathbf{v}_i^{\text{pref}}-\mathbf{v}'_i||
\end{equation}
\begin{equation}
    \mathbf{v}'_i=\arg \min_{\mathbf{v}''_i\in AV^i} penalty_i(\mathbf{v}''_i)
\end{equation}
We use $w_i=0.2$ for all robots, and set $\mathbf{v}_i^{\text{pref}}$ as the velocity of maximum robot speed, with heading directed towards its goal. $tc_i(\mathbf{v}'_i)$ is the expected time to collision with the robot's combined reciprocal velocity obstacle, and $AV^i$ is the set of admissible velocities.
Similar to our application of APFs, we choose the angular velocity that results in the closest alignment with vector $\mathbf{v}'_i$ in a single control step, and we choose the linear acceleration leading to the speed closest to that of $\mathbf{v}'_i$ in a single control step, as the control action.
The RVO agent of our work is based on the implementation in \cite{Guo2023multirobot}, which can be conveniently integrated into our simulation environment used in this study.

Our experimental results are summarized in Fig. \ref{fig:exp}. 
Adaptive IQN achieves the highest success rates across different levels of environment complexity, while consuming nearly the minimum amount of travel time and energy needed to reach goals.
Fig. \ref{fig:distributions} and Fig. \ref{fig:trajectories} visualize an experiment episode where seven robots, eight vortices and eight static obstacles exist.
In the left plot of Fig. \ref{fig:distributions}, robot 2 faces several static obstacles on the left and three robots coming from the right, while its goal is basically in the direction of its current speed.
With a higher risk sensitivity, the adaptive IQN agent chooses to turn left to avoid getting into a more congested state with higher collision risk, while the greedy IQN agent would keep moving in the current direction towards the goal.
Subsequently, after robot 2 turns left, robot 3 faces head-on collision risk with robot 2, and the adaptive IQN agent directs robot 3 to turn left in this instance to get away from the head-on situation.
In comparison, Fig. \ref{fig:trajectories} shows that greedy IQN failed in this episode due to the collision between robot 0 and robot 2.
Thus, adaptive IQN can effectively enhance the safety in congested environments, which is shown as the increase in the difference of success rate between adaptive IQN and greedy IQN, while keeping the travel time and energy consumption small. 
It can be seen from Fig. \ref{fig:exp} that DQN requires the minimum travel time and energy cost in its successful episodes, and its planned trajectories are straight compared to that of other methods.
This evidence indicates the greedy nature of DQN, the success rate of which drops significantly as the environment becomes more complex.
APF and RVO are vulnerable to current disturbances of robot motions, which cause the ASVs to struggle near vortices, and to fall victim to collisions with static obstacles in Fig. \ref{fig:trajectories}. 

\vspace{-2mm}
\section{Conclusion}
\label{sec:conclusion}


We propose a Distributional RL based decentralized multi-ASV collision avoidance policy, which is deployed in simulated congested marine environments filled with unknown static obstacles and vortical current flows (inspired by the challenges of whitewater rafting).
The interactions among decentralized decision making agents as well as with static obstacles are considered in the policy network, which is shared by all agents during training to induce the coordinated avoidance of collisions, and robustness to current disturbances.  
Compared to traditional DRL, APF and RVO, the Distributional RL policy dominates in navigation safety, which can be further enhanced by adapting the risk sensitivity, while also achieving nearly the minimum amount of travel time and energy consumption.

\vspace{-2mm}
\section*{ACKNOWLEDGMENT}

This research was supported by the Office of Naval Research, grants N00014-20-1-2570 and  N00014-21-1-2161.

\bibliographystyle{IEEEtran}
\bibliography{main}

\end{document}